\pdfoutput=1
\documentclass[11pt]{article}
\usepackage[preprint]{acl}
\usepackage{times}
\usepackage{latexsym}
\usepackage[T1]{fontenc}
\usepackage[utf8]{inputenc}
\usepackage{microtype}
\usepackage{inconsolata}
\usepackage{graphicx}
\usepackage{booktabs}       
\usepackage{amsfonts}       
\usepackage{nicefrac}       
\usepackage{microtype}      
\usepackage{xcolor}         
\usepackage{graphicx}
\usepackage{amsmath}
\usepackage{wrapfig}
\usepackage{subcaption}
\usepackage{tcolorbox}
\tcbuselibrary{breakable} 
\usepackage{enumitem}
\usepackage{marvosym}
\definecolor{myblue}{RGB}{150, 150, 230}
\newcommand\music{\raisebox{-9pt}{\includegraphics[width=2em]{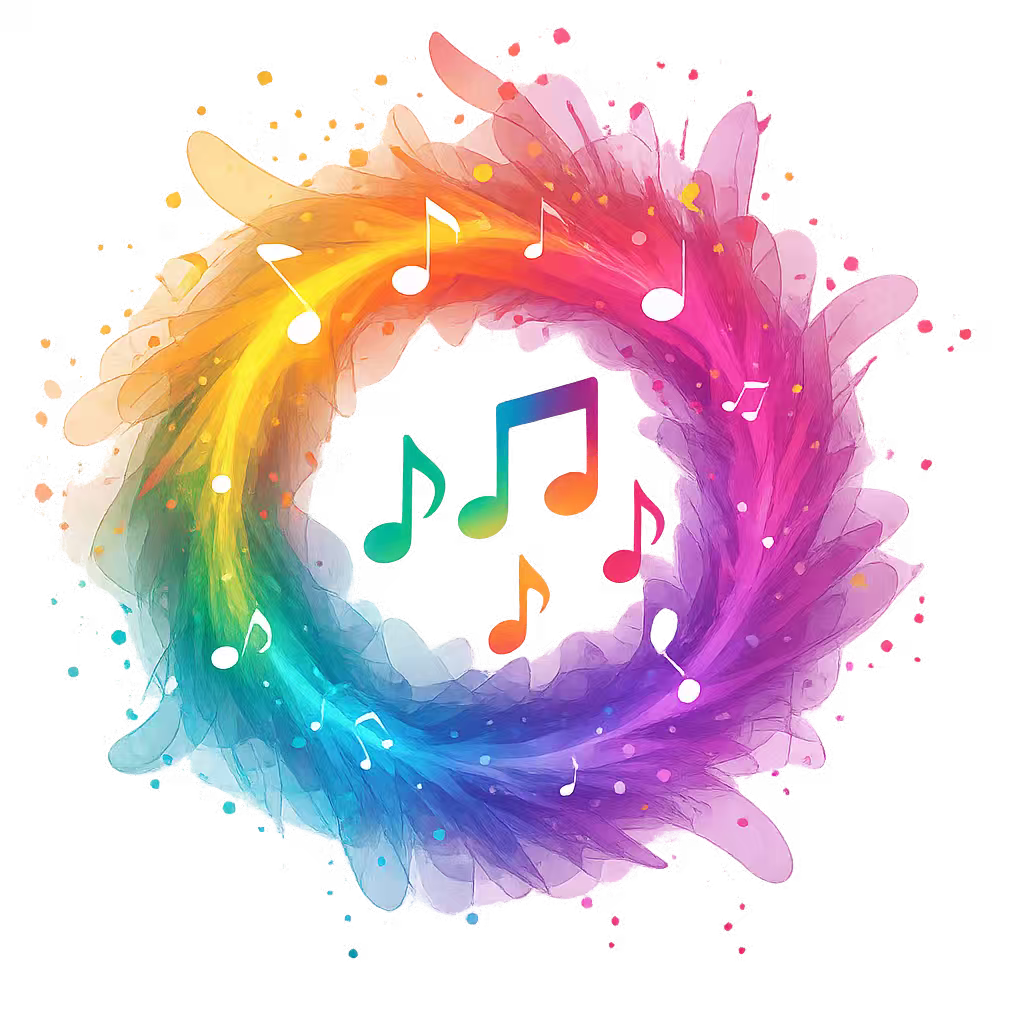}}}

\newcommand{\ie}{\textit{i.e.}}
\newcommand{\eg}{\textit{e.g.}}
\newcommand{\vs}{\textit{v.s.}}

\title{\music{} Multimodal Large Language Models for \underline{MU}lti-\underline{S}ubject \underline{I}n-\underline{C}ontext Image Generation}

\author{Yucheng Zhou, ~Dubing Chen, ~Huan Zheng, ~Jianbing Shen$^{\text{\Letter}}$\\
SKL-IOTSC, CIS, University of Macau \\
\texttt{yucheng.zhou@connect.um.edu.mo, jianbingshen@um.edu.mo}
}

\begin{document}
\maketitle
\renewcommand{\thefootnote}{\Letter} 
\footnotetext{Corresponding Author.}

\begin{abstract}
Recent advances in text-to-image (T2I) generation have enabled visually coherent image synthesis from descriptions, but generating images containing multiple given subjects remains challenging. As the number of reference identities increases, existing methods often suffer from subject missing and semantic drift. To address this problem, we propose MUSIC, the first MLLM specifically designed for \textbf{MU}lti-\textbf{S}ubject \textbf{I}n-\textbf{C}ontext image generation. To overcome the data scarcity, we introduce an automatic and scalable data generation pipeline that eliminates the need for manual annotation. Furthermore, we enhance the model's understanding of multi-subject semantic relationships through a vision chain-of-thought (CoT) mechanism, guiding step-by-step reasoning from subject images to semantics and generation. To mitigate identity entanglement and manage visual complexity, we develop a novel semantics-driven spatial layout planning method and demonstrate its test-time scalability. By incorporating complex subject images during training, we improve the model's capacity for chained reasoning. In addition, we curate MSIC, a new benchmark tailored for multi-subject in-context generation. 
Experimental results demonstrate that MUSIC significantly surpasses other methods in both multi- and single-subject scenarios.
\end{abstract}

\section{Introduction}
Recent years have witnessed significant advancements in image generation technologies, particularly in text-to-image (T2I) generation~\cite{zhang2023adding,flux2024,esser2024scaling,podell2023sdxl,xiao2024omnigen,yu2022scaling,zhou2026less}, where models have shown enhanced capabilities in semantic understanding and image synthesis. With the increasing demand for personalized applications, a growing body of research has focused on personalized image generation~\cite{zhang2023adding,mou2024t2i,gal2022image,ruiz2023dreambooth,hu2022lora}. These methods incorporate reference images or subject identities into the generation process to ensure alignment with the input text while preserving the visual identity of the reference subjects. This task holds practical relevance for applications such as multi-person scene synthesis and complex product visualization.

\begin{figure}[t]
    \centering
    \includegraphics[width=\linewidth]{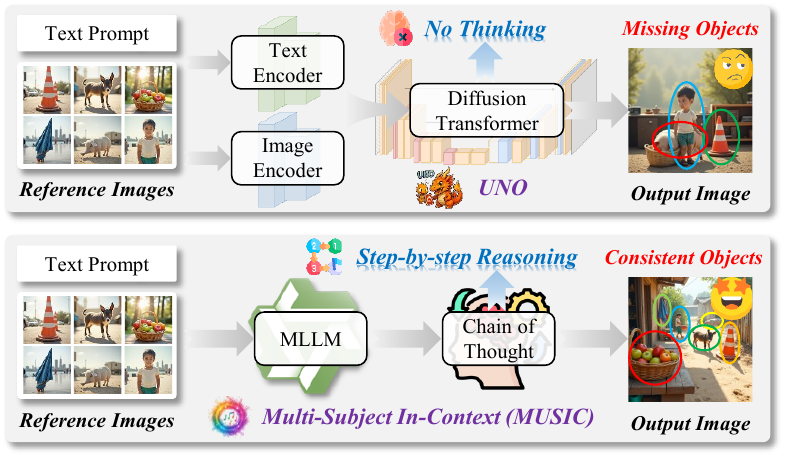}
    \includegraphics[width=\linewidth]{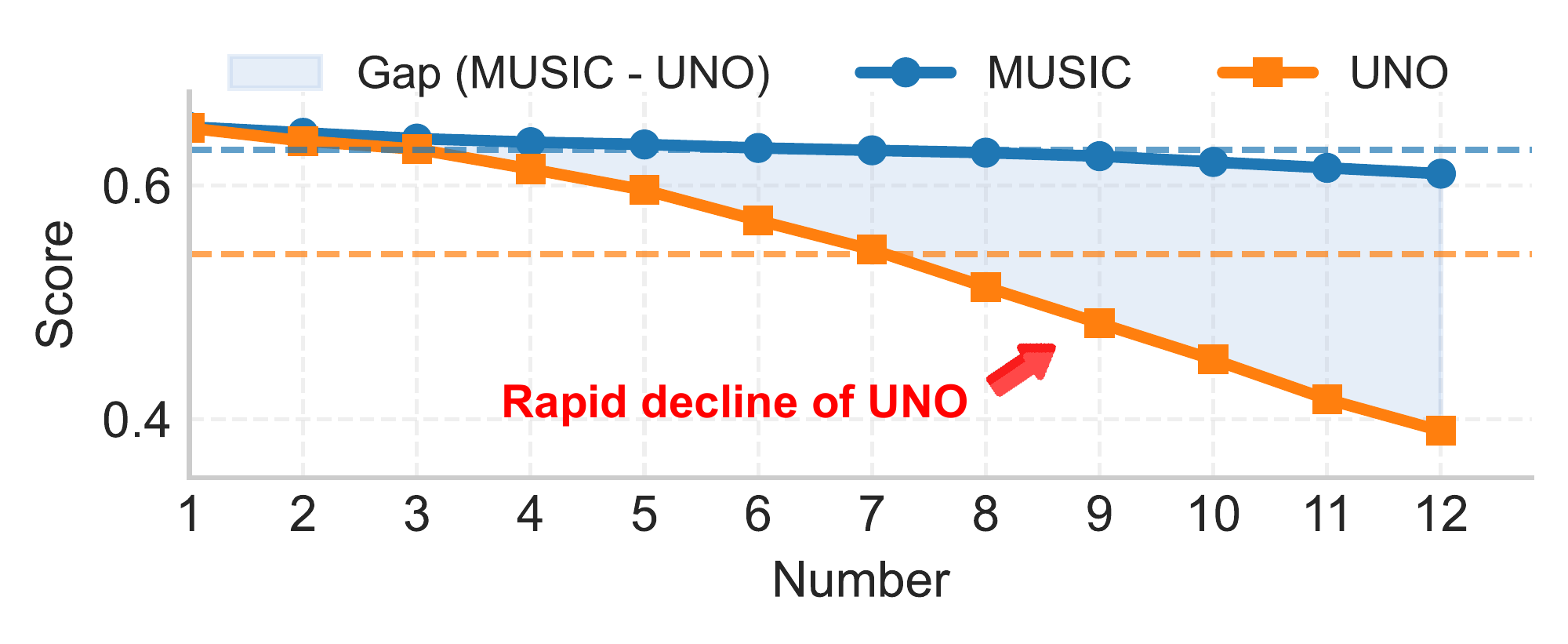}
    \vspace{-8mm}
    \caption{\small \textbf{Top:} Comparison of our MUSIC (bottom) \vs the subject-to-image method UNO (top). \textbf{Bottom:} UNO struggles as the number of subject images grows. Our MLLM-based method uses a thinking mechanism and more effectively generates the scene with multi-subjects.}
    \label{fig:intro}
    \vspace{-3mm}
\end{figure}

Existing approaches face substantial performance degradation as the number of reference subjects increases, as shown in Figure~\ref{fig:intro} (Right). Challenges such as subject missing and semantic drift frequently emerge, revealing the scalability limitations of current methods in complex, multi-subject generation scenarios. In contrast to diffusion models (\eg, DiT~\cite{peebles2023scalable}), multimodal large language models (MLLMs)~\cite{ge2023making,ge2024seed,zhou2024rethinking} exhibit superior generalization and contextual reasoning capabilities in both vision understanding and language reasoning. It has the potential to handle more complex instructions and intricate semantic relationships across multiple subjects. This raises a critical question: \textit{``how can we effectively harness the reasoning power of MLLMs to enhance the quality and consistency of multi-subject in-context image generation?''}

To address this challenge, we propose a novel MUSIC model, the first MLLM tailored for \textbf{MU}lti-\textbf{S}ubject \textbf{I}n-\textbf{C}ontext image generation. 
We design an automatic data generation pipeline that eliminates the need for manual annotation. 
As shown in Figure~\ref{fig:intro} (Left), we introduce a vision chain-of-thought (CoT)  that guides the MLLM to perform step-by-step reasoning from subject images to semantic modeling and final image generation. 
This approach substantially improves the model's ability to understand multi-subject semantic relationships.
To handle the visual complexity and mitigate identity entanglement in multi-subject images, we propose a novel semantics-driven spatial layout planning method that assigns semantic ownership to visual content before generation, thereby reducing semantic conflicts. 
We also explore the test-time scalability potential of this planning method.
Furthermore, by incorporating complex subject images in our model training, we enhance the model's capacity for chained reasoning across subject images, semantics, and generated images.

To better evaluate multi-subject image generation, we manually curated a new benchmark dataset, MSIC, from synthetic data, specifically designed for this task. Experimental results on the MSIC dataset and DreamBench~\cite{ruiz2023dreambooth} demonstrate that our method significantly outperforms existing approaches in semantic consistency and identity fidelity in both multi- and single-subject image generation. Our main contributions are summarized as below:

\begin{itemize}[leftmargin=*]
\item We propose \textbf{MUSIC}, the first MLLM designed for multi-subject in-context image generation, integrating vision reasoning capabilities for multi-subject understanding and generation.
\item We design an automatic and scalable data generation pipeline that requires no raw data, enabling efficient large-scale training data construction.
\item We enhance the model's multi-subject understanding through a vision chain-of-thought mechanism, semantics-driven spatial layout planning, and training on complex subject images. We verify the scalability of layout planning at test time.
\item We introduce the MSIC benchmark for multi-subject in-context image generation, and our method achieves state-of-the-art performance on MSIC and DreamBench.
\end{itemize}

\section{Related Work}\label{app:related}
\subsection{Generative Models}

\paragraph{Diffusion Models.}
Diffusion models~\cite{sohl2015deep,ho2020denoising,dhariwal2021diffusion,song2020denoising,peebles2023scalable} generate high-fidelity images via a forward noising and reverse denoising process. DDPM~\cite{ho2020denoising} first achieved competitive image quality, and subsequent work surpassed GANs~\cite{goodfellow2020generative,brock2018large,karras2020analyzing} in sample quality~\cite{dhariwal2021diffusion}. LDMs~\cite{rombach2022high} reduce computational costs by operating in latent space, while DDIM~\cite{song2020denoising} and consistency models~\cite{song2023consistency} accelerate sampling. Scalable transformer architectures such as DiT~\cite{peebles2023scalable} and U-ViT~\cite{bao2023all} further boost performance. Recent advances include architectural improvements via long-skip-connections with spectral constraints~\cite{chen2025towards}, hierarchical compositional generation~\cite{yang2025hicogen}, and efficient sampling strategies~\cite{wang2026ladr}. For controllable generation, DC-ControlNet~\cite{yang2025dc} decouples inter- and intra-element conditions, while self-rewarding LVLMs~\cite{yang2025self} optimize T2I prompts via model-driven feedback.

\paragraph{Autoregressive Generative Models.}
Autoregressive models generate images by modeling pixel or token sequences, inspired by language modeling~\cite{brown2020language,achiam2023gpt,touvron2023llama,qwen2.5}. Early models like PixelRNN~\cite{van2016pixel} and PixelCNN~\cite{van2016conditional} predicted pixels sequentially with low efficiency. Modern approaches use vector-quantized representations such as VQ-VAE~\cite{razavi2019generating} and VQ-GAN~\cite{esser2021taming} to compress images into discrete tokens~\cite{van2017neural}. Masked methods including MaskGIT~\cite{chang2022maskgit}, Muse~\cite{chang2023muse}, and MaskBit~\cite{weber2024maskbit} enable parallelized generation via masked token prediction~\cite{gao2023masked,fan2024fluid}. Recent models like LLaMAGen~\cite{sun2024autoregressive}, Show-o~\cite{xie2024show}, Infinity~\cite{han2024infinity}, and Emu3~\cite{wang2024emu3} scale up decoder-only autoregressive frameworks~\cite{achiam2023gpt,touvron2023llama,qwen2.5}. VAR~\cite{tian2024visual} redefines generation as next-scale prediction, Fluid~\cite{fan2024fluid} integrates continuous tokenization with diffusion loss, and RandAR~\cite{pang2024randar} enables parallel decoding via position prediction with KV-Cache~\cite{pope2023efficiently}. More recently, vision representation compression has improved autoregressive video generation efficiency~\cite{zhou2026less}, diffusion loss has been incorporated to refine condition errors~\cite{zhoucondition}, and entropy-guided optimization balances exploration and stability~\cite{songbroad}.

\subsection{In-Context Learning.}
\paragraph{LLM In-Context Learning}
In-context learning (ICL) enables LLMs to adapt to new tasks via few-shot prompts without parameter updates~\citep{brown2020language}. Subsequent studies explored ICL mechanisms, framing it as implicit Bayesian inference~\cite{xie2021explanation} and identifying attention-driven task vectors~\cite{olsson2022context}. Chain-of-Thought prompting~\citep{wei2022chain} significantly boosts reasoning, while retrieval-based example selection~\citep{liu2021makes}, context calibration~\citep{zhao2021calibrate}, and example ordering~\citep{lu2021fantastically} further optimize ICL performance. Recent work shows that weak models can elicit strong performance through multi-capability supervision~\cite{zhou2025weak}.

\paragraph{Vision In-Context Learning.}
Vision in-context learning extends ICL to visual tasks using few visual examples~\citep{tsimpoukelli2021multimodal}. Flamingo~\citep{alayrac2022flamingo} leverages interleaved image-text prompts for few-shot visual QA, while visual prompting via inpainting~\citep{bar2022visual} specifies tasks through image patches. Recent work demonstrates improved compositional understanding through ICL~\citep{nulli2024context} and effective visual task adaptation in LVLMs via in-context examples~\citep{zhou2024visual}. Challenges remain in efficiently representing visual prompts and generalizing with limited context~\citep{zhang2023makes,zhou2024rethinking}.

\subsection{Subject-Driven Image Generation}
Subject-driven image generation synthesizes specific subjects in novel contexts. Early approaches~\cite{zhang2023adding,mou2024t2i,gal2022image,ruiz2023dreambooth,hu2022lora} require per-subject optimization, limiting generalization. IP-Adapter~\citep{ye2023ip}, BLIP Diffusion~\citep{li2023blip}, and ELITE~\cite{wei2023elite} achieve zero-shot generation via additional image encoders. Recent methods tackle multi-subject scenarios~\cite{ma2024subject,huang2025resolving,wang2024ms,wu2025less}: MS-Diffusion~\cite{wang2024ms} uses layout guidance, UNO~\cite{wu2025less} employs vision in-context learning for flexible multi-subject synthesis, and unified multimodal agents~\cite{chen2026unify} combine diverse capabilities for world-grounded image synthesis.

Despite these advances, most existing approaches struggle to generalize to large-scale multi-subject combinations, and constructing diverse, high-quality training datasets for multi-subject generation remains challenging. This work addresses these gaps through a fully automated data construction pipeline and enhanced identity consistency via vision in-context learning, enabling robust multi-subject generation.

\begin{figure*}[!t]
    \centering
    \includegraphics[width=\textwidth]{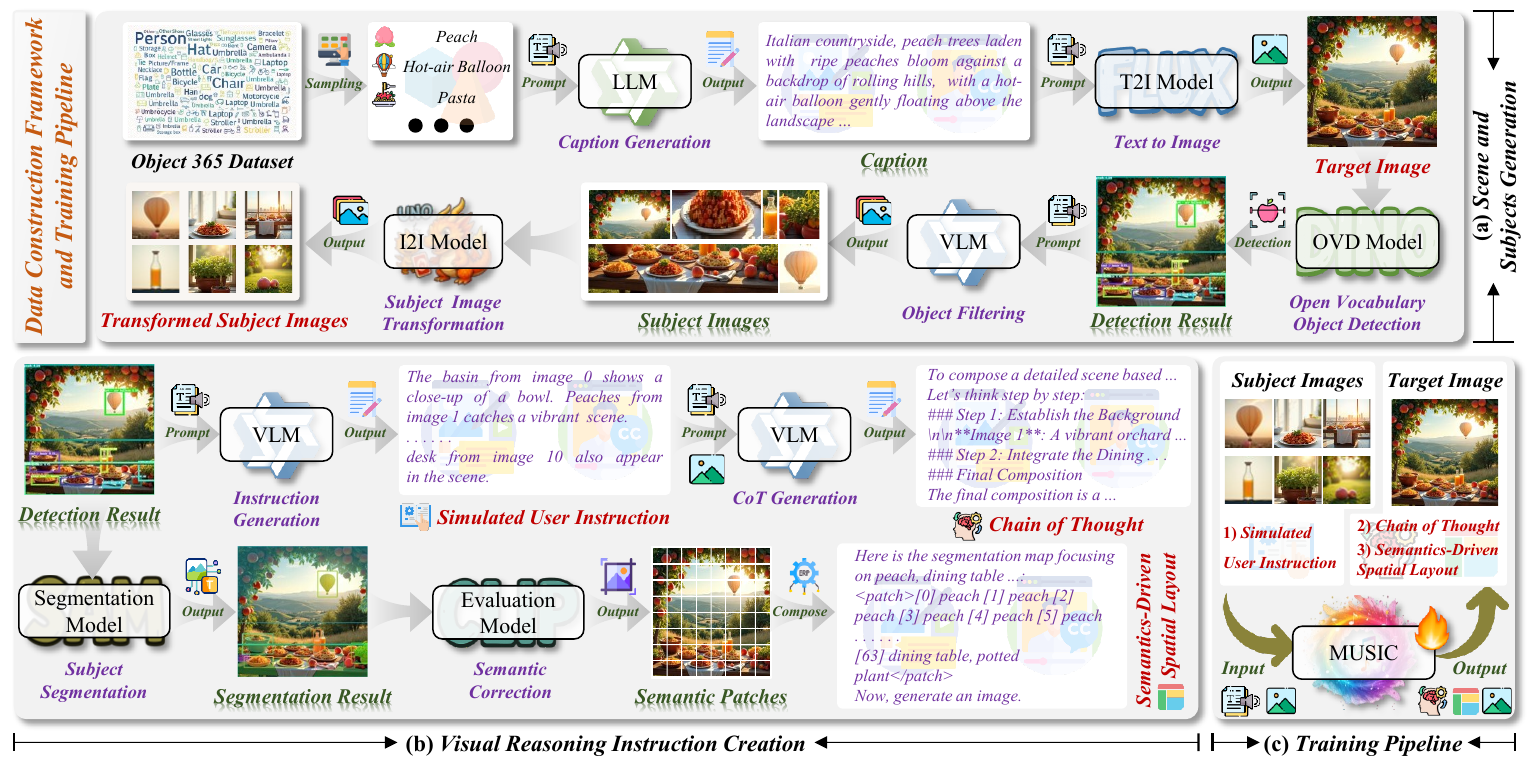}
    \vspace{-8mm}
    \caption{\small \textbf{Overview of our data construction pipeline:} \textbf{(a)} An LLM and T2I model generate the target image, followed by an OVD model detecting subjects. A VLM filters out unsuitable objects, and an I2I model creates transformed subject images. \textbf{(b)} A VLM produces simulated user instructions and CoT instructions. A segmentation model generates segmentation masks, yielding a semantics-driven spatial layout textual description. \textbf{(c)} Application of the constructed data pair in the training pipeline.}
    \label{fig:pipeline}
    \vspace{-3mm}
\end{figure*}

\section{Methodology}
To address the data scarcity for multi-subject in-context generation, we introduce an automated framework for generating diverse multi-subject datasets. We then detail the training of our MUSIC model for multi-subject in-context image generation and discuss strategies for test-time scaling using our semantics-driven spatial layout planning.

\subsection{Automated Multi-Subject Dataset Generation Framework}
\label{sec:multi_subject_data}
The core of our data generation strategy is a novel, fully automated framework for synthesizing multi-subject training data. Critically, this framework operates without relying on any pre-existing image or text metadata, enabling scalable dataset creation from foundational models.
Our pipeline integrates multiple foundation models, \ie, large language model (LLM)~\cite{qwen2.5}, text-to-image (T2I) model~\cite{blackforestlabs_flux}, vision language model (VLM)~\cite{qwen2.5}, image-to-image (I2I) model~\cite{wu2025less}, open-vocabulary detection (OVD) model~\cite{liu2024grounding}, and foundational segmentation model~\cite{ravi2024sam}, to produce subject-consistent image pairs, instructions, and reasoning sequences for training multi-subject in-context generation models. 
As shown in Figure~\ref{fig:pipeline}, the pipeline includes several stages: category sampling, caption generation, scene generation, subject detection and filtering, perspective transformation, vision chain-of-thought (CoT) reasoning, and semantics-driven spatial layout planning.

\subsubsection{Scene and Subjects Generation}
\paragraph{Subject Sampling and Scene Generation.}
As shown in Figure~\ref{fig:pipeline} (a), we start by random select between $2n_{\text{min}}$ and $2n_{\text{max}}$ categories from the Object365 dataset~\cite{shao2019objects365}. Here, $n_{\text{min}}$ and $n_{\text{max}}$ define the range of subject counts. Randomly sampled categories often lack semantic coherence. For example, ``lamp'' and ``pineapple'' are unlikely to appear naturally together. To address this, we use an LLM to reason over the candidate pool and automatically select a semantically related subset. The LLM then generates a caption describing a plausible scene that combines these categories.
This caption is fed into a T2I model to generate a target image $I_{\text{tgt}}$. This image serves as the foundation for the next stages and represents a complex multi-subject environment.

\paragraph{Subject Detection and Regeneration.}
To extract subjects in image $I_{\text{tgt}}$, we apply an OVD model to identify subjects in the target image and obtain their bounding boxes. To remove noisy detections, boxes with areas smaller than a threshold $\delta \cdot \text{Area}(I_{\text{tgt}})$ are filtered out. We then select up to $n_{\text{max}}$ subjects per image, prioritizing category diversity and favoring larger bounding boxes.

Since the OVD may produce incorrect detections, we further verify bounding boxes using a VLM. Specifically, we overlay the detected boxes, categories, and confidence scores on the original image and prompt the VLM to assess their correctness. Incorrect boxes are filtered out.
Each valid subject is cropped from the target image to create a subject-specific subimage. To avoid the subject-to-image generation model learning trivial copy-paste strategies, we apply an I2I model to produce view-transformed subject images. These transformed images $I_{\text{subj}_1}, I_{\text{subj}_2}, \dots, I_{\text{subj}_S}$ and the target image $I_{\text{tgt}}$ form the paired training data for our model, where $S$ is the number of detected subjects.

\subsubsection{Visual Reasoning Instruction Creation}
\paragraph{Instruction and Vision CoT Generation.}
In real scenarios, users typically provide a simple instruction along with a set of subject images to guide scene composition. However, as the number of subjects increases, maintaining subject identity and correct spatial arrangements becomes very challenging. To solve this, we generate CoT data to improve the model's reasoning abilities.

As shown in Figure~\ref{fig:pipeline} (b), we use a VLM to generate a simulated user instruction based on the target image annotated with bounding boxes and category labels. Each detected subject is assigned an ID to clarify object references. The instruction, along with the annotated target image, is then used to prompt another VLM to produce a detailed CoT reasoning sequence. The simulated user instruction may include subject IDs or omit them; we generate both types to cover a wide range of scenarios. Subjects mentioned in the CoT are expected to be linked to their corresponding IDs. This helps the model understand subject interactions within the scene, especially when multiple instances of the same category appear. We also perform string matching to ensure that IDs in the CoT and instruction align with detected subjects, removing any incorrect IDs introduced by hallucination. The CoT outlines a clear, step-by-step reasoning process for scene construction, including spatial arrangements and contextual relationships between subjects.

\paragraph{Semantics-Driven Spatial Layout Planning.}
To further enhance subject placement, we propose a semantics-driven spatial layout planning approach. Following vision CoT reasoning~\cite{shao2024visual}, this method aligns visual content more precisely within the 2D spatial domain of the image. Specifically, we incorporate category-level semantic information into designated spatial regions.
Inspired by Text4Seg~\cite{lan2024text4seg}, we partition the target image into an $M \times M$ grid ($M=8$ in our implementation) and determine the dominant subject categories for each patch.

To assign categories accurately, we use the foundational segmentation model SAM2~\cite{ravi2024sam} to generate object masks from detected bounding boxes. However, as SAM2 relies solely on spatial inputs and lacks language understanding, its masks may miss the correct regions. To improve robustness, we dynamically choose between using the mask or the bounding box for each subject. Specifically, we employ CLIP~\cite{radford2021learning} as the evaluation model to compute cosine similarities between the class text embedding and the average visual features extracted from both the masked and unmasked regions:
\begin{align}
\text{Sim}_{\text{mask}} &= \cos\left( f_{\text{mask}}, f_{\text{class}} \right), \notag\\
\text{Sim}_{\text{unmask}} &= \cos\left( f_{\text{unmask}}, f_{\text{class}} \right).
\end{align}
If $\text{Sim}_{\text{mask}} > \text{Sim}_{\text{unmask}}$, the mask is used, otherwise, we fall back to the bounding box. This approach balances SAM2's fine-grained segmentation with improved robustness.

We compute the intersection-over-union (IoU) between each patch and the subject's mask or bounding box. To avoid filtering out small objects, we use a dynamic IoU threshold:
\begin{align}
\tau = \lambda \cdot \frac{1}{K} \sum_{i=1}^{K} \text{IoU}(b_i, p_i),
\end{align}
where $\lambda$ is the pre-defined scaling factor, $K$ is the number of non-zero IoUs, and $b_i$ and $p_i$ represent the subject region and the patch, respectively.
Patches with IoU larger than $\tau$ are assigned the subject's category; otherwise, they default to ``others.'' The final spatial layout prompt is structured as follows:

\begin{verbatim}
Here is the segmentation map focusing 
on flower, frame, pineapple, plate,
potted plant, lamp, couch: <patch> [0] 
others [1] others [2] frame [3] others 
... [63] others </patch>
Now, generate an image.
\end{verbatim}

This segmentation-aware instruction is integrated into the CoT to guide the subject-to-image generation model in accurately composing multi-subject scenes with correct spatial layouts.

\subsubsection{Final Training Data Construction}
We further enriched our dataset by simulating more realistic user demands, such as generating scenes centered around a specific subject within a complex environment. To achieve this, we used an LLM to construct a category dictionary, where each key represents a class and the corresponding value contains semantically similar classes. When generating transformed subject images with the I2I model, we randomly select multiple similar classes from this dictionary and incorporate them into the instruction. This encourages the I2I model to produce more diverse and complex scenes. As a result, our dataset contains both simple and complex instances, providing a comprehensive training resource. More details can be found in Appendix~\ref{app:complex}.

We further augment the generated data by systematically reducing the number of subjects. We sort subjects by bounding box size and iteratively remove the smallest ones until only two subjects remain. Correspondingly, we update the instruction by removing or reordering subject IDs to match the new subject set. Through this process, we generate data with varying subject counts at no extra cost, helping the model better handle scenes with different levels of complexity.

As shown in Figure~\ref{fig:pipeline} (c), each final training instance includes a set of augmented subject images $\{I_{\text{subj}_1}, \dots, I_{\text{subj}_S}\}$, the target image~$I_{\text{tgt}}$, a concise simulated user instruction~$T_{\text{instr}}$, a detailed CoT reasoning sequence $C_{\text{CoT}}$, and a semantics-driven spatial layout planning description $L_{\text{spatial}}$. This rich and structured supervision significantly improves the model's ability to reason, compose, and maintain both subject identity and spatial consistency in complex multi-subject scenarios.

\subsection{Training MUSIC for Multi-Subject In-Context Image Generation}
\label{sec:training_music}

We train our proposed model, MUSIC (Multi-Subject In-Context Image Generation), using the diverse and structured datasets automatically generated by the framework described in Section \ref{sec:multi_subject_data}. The architecture of MUSIC is built upon the principles of existing MLLM frameworks, \eg, SEED-X~\cite{ge2024seed}, which are adept at processing and generating both textual and visual data. Our training strategy decomposes the complex task of multi-subject in-context generation into two distinct capabilities, mirroring the structure of our generated data and the desired inference process:

\paragraph{Capability 1: Visual Reasoning and Spatial Planning.}
This phase focuses on teaching the model to interpret user intent and plan the scene composition. The input consists of the initial user instruction (text) and the set of augmented subject images $\{I_{\text{subj}_1}, \dots, I_{\text{subj}_S}\}$. The training objective is to predict the corresponding vision CoT reasoning sequence and the semantics-driven spatial layout plan generated by our pipeline.

Formally, given the input instruction $T_{\text{instr}}$ and subject images $\{I_{\text{subj}_i}\}_{i=1}^S$, this capability learns a mapping $f_1$:
\begin{align}
f_1(T_{\text{instr}}, \{I_{\text{subj}_i}\}_{i=1}^S) \rightarrow (\hat{C}_{\text{CoT}}, \hat{L}_{\text{spatial}})
\end{align}
where $\hat{C}_{\text{CoT}}$ is the predicted CoT text sequence and $\hat{L}_{\text{spatial}}$ is the predicted spatial layout grid. This capability is trained using supervision from the ground-truth $C_{\text{CoT}}$ and $L_{\text{spatial}}$, with a cross-entropy loss. The loss for this stage is denoted as $L_1$.

\paragraph{Capability 2: Image Generation from Plan.}
This phase focuses on synthesizing the final image conditioned on the outputs of the planning stage. The input to this capability consists of the vision CoT and the semantics-driven spatial layout plan. The training objective is to generate the target image $I_{\text{tgt}}$ that reflects the intended composition and semantics.
This capability learns a mapping $f_2$:
\begin{align}
f_2(C_{\text{CoT}}, L_{\text{spatial}}) \rightarrow \hat{I}_{\text{tgt}}
\end{align}
where $\hat{I}_{\text{tgt}}$ denotes the synthesized image. During training, we leverage the ground-truth $C_{\text{CoT}}$ and $L_{\text{spatial}}$ as conditioning signals. Following the SEED-X framework, this stage is optimized by minimizing the distance between the visual representations of the synthesized image $\hat{I}_{\text{tgt}}$ and the ground-truth image $I_{\text{tgt}}$. The corresponding training loss is denoted as $L_2$.

\paragraph{Overall Training Objective.}
The MUSIC model is trained by optimizing a combination of the losses from both capabilities. The total loss $L$ can be expressed as:
\begin{align}
L &= w_1 L_1(T_{\text{instr}}, \{I_{\text{subj}_i}\}_{i=1}^S; C_{\text{CoT}}, L_{\text{spatial}}) \notag\\
&+ w_2 L_2(C_{\text{CoT}}, L_{\text{spatial}}; I_{\text{tgt}})
\end{align}
where $w_1$ and $w_2$ are weighting factors. Training is performed end-to-end, allowing the gradients from $L_2$ to flow back by the predicted CoT and layout.

\subsection{Test Time Scaling via Semantics-Driven Spatial Layout Planning}
\label{sec:test_time_scaling}
At inference time, we enable scalable and diverse generation by producing multiple candidate spatial layout plans from a single instruction and subject set. MUSIC first generates $N$ candidate planning branches, each comprising a unique CoT and spatial layout:
\begin{align}
\{(\hat{C}_{\text{CoT}}^{(j)}, \hat{L}_{\text{spatial}}^{(j)})\}_{j=1}^N
\end{align}

Each plan leads to a synthesized image via the generation module:
\begin{align}
\hat{I}_{\text{tgt}}^{(j)} = f_2(\hat{C}_{\text{CoT}}^{(j)}, \hat{L}_{\text{spatial}}^{(j)})
\end{align}

To select the best output, we use a CLIP-based verifier to compute similarity scores between each generated image and the original instruction $T_{\text{instr}}$:
\begin{align}
I_{\text{final}} &= \hat{I}_{\text{tgt}}^{(j^*)}, \\
j^* &= \arg\max_{j} \cos\left(f_{\text{CLIP}}^{\text{image}}(\hat{I}_{\text{tgt}}^{(j)}), f_{\text{CLIP}}^{\text{text}}(T_{\text{instr}})\right)\notag
\end{align}

This strategy introduces minimal overhead while significantly enhancing output diversity and fidelity. It is adaptable to different generation frameworks and supports flexible test-time control over layout complexity.

\section{Experiments}

\subsection{Experiments Setting}
\paragraph{Implementation Details.}\label{sec:setting}
We adopt Qwen-3~\cite{qwen2.5} as our LLM, Qwen-2.5 VL~\cite{qwen2.5} as our VLM, FLUX-1.0-DEV~\cite{flux2024} as the T2I model, UNO-FLUX-1.0-DEV~\cite{wu2025less} for I2I generation, GroundingDINO as our OVD model, SAM2 for segmentation, and CLIP-vit-large-patch14~\cite{radford2021learning} for mask evaluation. We set the maximum number of subjects $n_{max}$ to $12$, the minimum number $n_{min}$ to $1$, the threshould scaling factor $\delta$ to 0.01, the patch count to $8\times8$, and the scaling factor $\lambda$ to $0.05$.
We initialize our model using SEED-X and apply Low-Rank Adaptation (LoRA) for efficient fine-tuning. The training is conducted on a synthetic dataset comprising 10{,}000 samples generated through our customized data generation pipeline. All experiments are performed using a server equipped with 8~$\times$~A100 GPUs. LoRA is configured with a rank of 64 and a scaling factor of $\alpha = 64$. We train the model for 10 epochs using a learning rate of $1 \times 10^{-4}$. The LoRA weight scaling parameters are fixed at $w_1 = 0.5$ and $w_2 = 0.5$ throughout the training process.

\paragraph{Evaluation Benchmarks and Metrics.}
We evaluate our proposed MUSIC framework on two benchmark datasets targeting different aspects of in-context image generation. For multi-subject evaluation, we introduce the MSIC dataset, designed to assess model scalability across varying subject counts (1 to 12). For single-subject generation, we adopt DreamBench~\cite{ruiz2023dreambooth}. Performance is quantified using three automatic metrics: DINO~\cite{oquab2023dinov2}, CLIP-I, and CLIP-T~\cite{radford2021learning}. DINO measures image-level fidelity using self-supervised features. CLIP-I evaluates subject fidelity by comparing the generated image to the reference. CLIP-T measures alignment with the input text prompt. Higher scores indicate better quality for all metrics.

\begin{table}[t]\small
\centering
\setlength{\tabcolsep}{2pt}
\resizebox{\linewidth}{!}{
\begin{tabular}{lccc}
\toprule
\bf Method & \bf DINO ↑ & \bf CLIP-I ↑ & \bf CLIP-T ↑ \\
\midrule
Oracle (caption) & - & - & 0.339 \\
\midrule
Subject Diffusion~\cite{ma2024subject} & 0.513 & 0.702 & 0.287 \\
MIP-Adapter~\cite{huang2025resolving} & 0.497 & 0.715 & 0.288 \\
MS-Diffusion~\cite{wang2024ms} & 0.532 & 0.714 & 0.290 \\
OmniGen~\cite{xiao2024omnigen} & 0.525 & 0.714 & 0.300 \\
UNO~\cite{wu2025less} & 0.541 & 0.721 & 0.296 \\\midrule
MUSIC (Ours) & 0.622 & 0.812 & 0.322 \\
MUSIC* (Ours) & \bf 0.631 & \bf 0.822 & \bf 0.330 \\
\bottomrule
\end{tabular}}
\vspace{-3mm}
\caption{\small Performance comparison on the MSIC (multi-subject in-context generation).
MUSIC* refers to our method with test-time scaling. Oracle uses ground-truth captions.}
\label{tab:multi_comparison}
\vspace{-3mm}
\end{table}

\paragraph{Comparative Methods.}
To comprehensively assess the effectiveness of MUSIC, we compare it against a range of state-of-the-art baselines across both multi-subject and single-subject in-context image generation tasks. For the multi-subject setting, we include Subject Diffusion~\cite{ma2024subject}, MIP-Adapter~\cite{huang2025resolving}, MS-Diffusion~\cite{wang2024ms}, OmniGen~\cite{xiao2024omnigen}, and UNO~\cite{wu2025less} as comparison baselines. 
For single-subject setting, we include Textual Inversion~\cite{gal2022image}, DreamBooth~\cite{ruiz2023dreambooth}, BLIP-Diffusion~\cite{li2023blip}, ELITE~\cite{wei2023elite}, Re-Imagen~\cite{chen2022re}, BootPIG~\cite{purushwalkam2024bootpig}, SSR-Encoder~\cite{zhang2024ssr}, RealCustom++~\cite{mao2024realcustom++}, OmniGen~\cite{xiao2024omnigen}, OminiControl~\cite{tan2024ominicontrol}, FLUX.1 IP-Adapter, and UNO~\cite{wu2025less}. 

\subsection{Quantitative Evaluations}
\paragraph{Automatic Scores.}
\begin{table}[t]\small
\centering
\setlength{\tabcolsep}{2pt}
\resizebox{\linewidth}{!}{
\begin{tabular}{lccc}
\toprule
\textbf{Method} & \textbf{DINO ↑} & \textbf{CLIP-I ↑} & \textbf{CLIP-T ↑} \\
\midrule
Oracle (reference images) & 0.774 & 0.885 & - \\\midrule
Textual Inversion \cite{gal2022image} & 0.569 & 0.780 & 0.255 \\
DreamBooth \cite{ruiz2023dreambooth} & 0.668 & 0.803 & 0.305 \\
BLIP-Diffusion \cite{li2023blip} & 0.670 & 0.805 & 0.302 \\
ELITE \cite{wei2023elite} & 0.647 & 0.772 & 0.296 \\
Re-Imagen \cite{chen2022re} & 0.600 & 0.740 & 0.270 \\
BootPIG \cite{purushwalkam2024bootpig} & 0.674 & 0.797 & 0.311 \\
SSR-Encoder \cite{zhang2024ssr} & 0.612 & 0.821 & 0.308 \\
RealCustom++ \cite{mao2024realcustom++} & 0.702 & 0.794 & 0.318 \\
OmniGen \cite{xiao2024omnigen} & 0.693 & 0.801 & 0.315 \\
OminiControl \cite{tan2024ominicontrol} & 0.684 & 0.799 & 0.312 \\
FLUX.1 IP-Adapter \cite{flux-ipav2} & 0.582 & 0.820 & 0.288 \\
UNO \cite{wu2025less} & 0.760 & 0.835 & 0.304 \\\midrule
MUSIC (Ours) & 0.761 & 0.837 & 0.317  \\
MUSIC* (Ours) & \textbf{0.768} & \textbf{0.840} & \textbf{0.321}  \\
\bottomrule
\end{tabular}}
\vspace{-3mm}
\caption{\small Performance comparison of different methods on Dreambench (single-subject in-context generation).}
\label{tab:single_comparison}
\vspace{-4mm}
\end{table}
Table~\ref{tab:multi_comparison} presents results on the MSIC benchmark for multi-subject in-context generation. \textbf{MUSIC} outperforms all state-of-the-art baselines across three metrics (DINO 0.622, CLIP-I 0.812, CLIP-T 0.322), demonstrating superior identity preservation and semantic alignment in complex multi-subject prompts. The \textbf{MUSIC*} variant, with semantics-driven spatial planning for test-time scaling, achieves the best overall performance.
Table~\ref{tab:single_comparison} shows results on DreamBench~\cite{ruiz2023dreambooth} for single-subject generation. Although designed for multi-subject synthesis, \textbf{MUSIC} attains competitive results (DINO 0.761, CLIP-I 0.837) comparable to UNO~\cite{wu2025less} (0.760, 0.835), while \textbf{MUSIC*} slightly improves results (0.768, 0.840), confirming the benefit of test-time scaling.

\begin{figure}[!t]
    \centering
    \includegraphics[width=\linewidth]{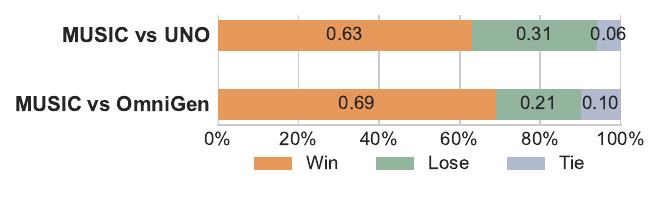}
    \vspace{-9mm}
    \caption{\small Human evaluation results comparing the MUSIC against OmniGen and UNO. }
    \label{fig:human_eval}
    \vspace{-3mm}
\end{figure}
\paragraph{Human Evaluation.}
To complement our automatic quantitative results, we conducted a human evaluation study comparing MUSIC against two strong baselines: OmniGen~\cite{xiao2024omnigen} and UNO~\cite{wu2025less}. For a subset of MSIC prompts, human raters compared paired images (MUSIC vs. baseline), evaluating overall quality, subject identity, and adherence to prompt instructions. Figure~\ref{fig:human_eval} presents the results. Against OmniGen, MUSIC was preferred in 69\% of cases vs. 21\% for OmniGen (10\% Tie). Against UNO, MUSIC was preferred in 63\% vs. 31\% for UNO (6\% Tie). These results demonstrate a strong human preference for MUSIC's outputs, aligning with our automatic metrics.

\subsection{Ablation Study}

\begin{figure}[!t]
    \centering
    \includegraphics[width=1\linewidth]{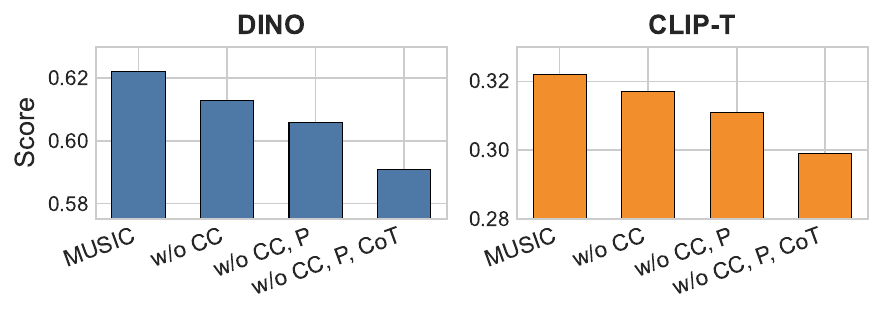}
    \vspace{-9mm}
    \caption{\small Ablation study. ``w/o CC'' (Complex Case data augmentation); ``w/o P'' (Complex Case and spatial layout planning); ``w/o CoT'' (Complex Case, spatial planning, and Chain-of-Thought reasoning).}
    \label{fig:ablation}
    \vspace{-3mm}
\end{figure}

As shown in Figure~\ref{fig:ablation}, an ablation study investigated the contribution of Complex Case augmentation (CC), spatial layout planning (P), and Vision Chain-of-Thought (CoT). Removing CC (``w/o CC'') caused a performance drop, suggesting its benefit for varying subject counts. A more significant decline, especially in CLIP-I and CLIP-T, occurred when P was also removed (``w/o CC, P''), highlighting its importance for accurate composition. The largest degradation was observed upon removing all three (``w/o CC, P, CoT''), underscoring CoT's critical role in guiding complex interactions and ensuring fidelity. The study confirms all components are beneficial, with P and CoT particularly impactful for reasoning in multi-subject generation.

\begin{figure}[!t]
    \centering
    \includegraphics[width=\linewidth]{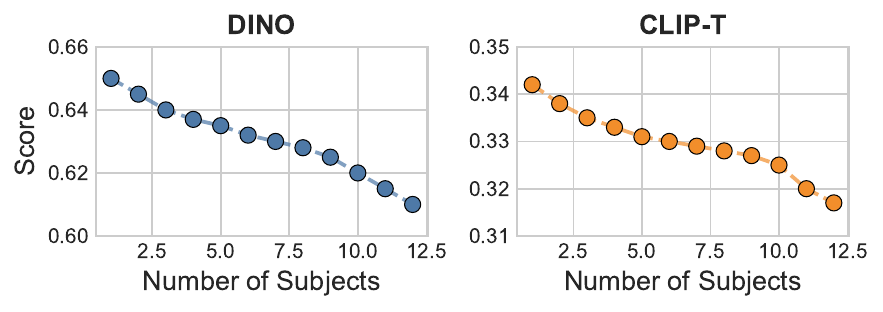}
    \vspace{-9mm}
    \caption{\small Performance of MUSIC as a function of the number of subjects on the MSIC.}
    \label{fig:subject_num}
    \vspace{-3mm}
\end{figure}

\begin{figure}[!t]
    \centering
    \includegraphics[width=\linewidth]{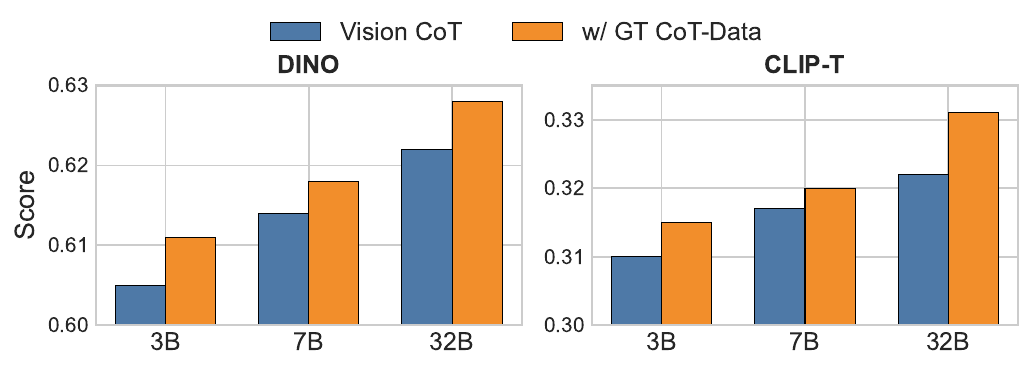}
    \vspace{-9mm}
    \caption{\small 
    Performance of MUSIC trained on Vision CoT data from Qwen2.5-VL (3B, 7B and 32B). ``w/ GT CoT-Data'' denotes inference with ground-truth CoT from Qwen2.5-VL, which further boosts generation quality.
    }
    \label{fig:cot}
    \vspace{-3mm}
\end{figure}

\begin{figure*}[t]
    \centering
    \includegraphics[width=1\linewidth]{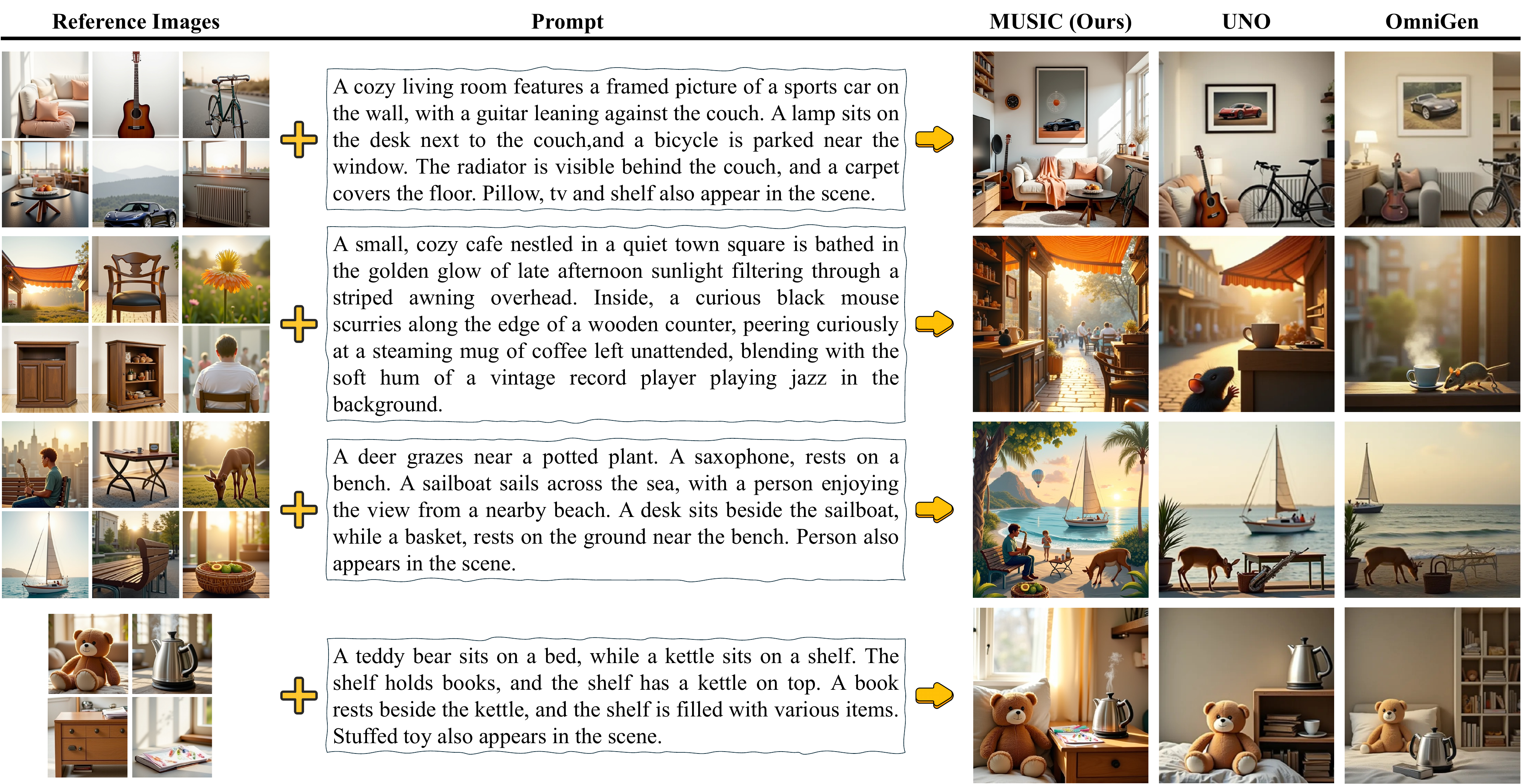}
    \vspace{-3mm}
    \caption{\small Qualitative comparison of multi-subject image generation results from \textbf{MUSIC (Ours)} against \textbf{UNO} and \textbf{OmniGen}. Each row shows the reference images, input prompt, and generated images.}
    \label{fig:case}
    \vspace{-1mm}
\end{figure*}

\paragraph{Effect of Subject Number.}
To assess MUSIC's performance under varying subject counts, we evaluated it on the MSIC with 1 to 12 subjects. As expected due to the inherent complexity of multi-subject scenes (maintaining identity, spatial relations, preventing entanglement), Figure~\ref{fig:subject_num} shows that DINO and CLIP-T scores decrease with the number of subjects. However, MUSIC maintains relatively strong performance even at high subject counts, demonstrating the robustness of our approach to multi-subject complexity.

\paragraph{Effect of Vision CoT.}
Vision Chain-of-Thought (CoT) reasoning is crucial for complex multi-subject scene understanding. We evaluated its impact using CoT data from varying Qwen2.5-VL model sizes (3B, 7B, 32B), reflecting VLM reasoning quality. Figure~\ref{fig:cot} shows that training with CoT from larger VLMs (32B) improves performance (DINO, CLIP-T scores) (left plots). Providing "ground-truth" CoT (from the 32B VLM) at inference significantly boosts performance across all models, regardless of training CoT quality (right plots). This highlights Vision CoT's vitality, the impact of training data quality, and the significant benefit of high-quality CoT during inference for complex multi-subject generation.

\subsection{Semantics-Driven Spatial Layout Planning for Test-Time Scaling}

\begin{figure}[t]
    \centering
    \includegraphics[width=\linewidth]{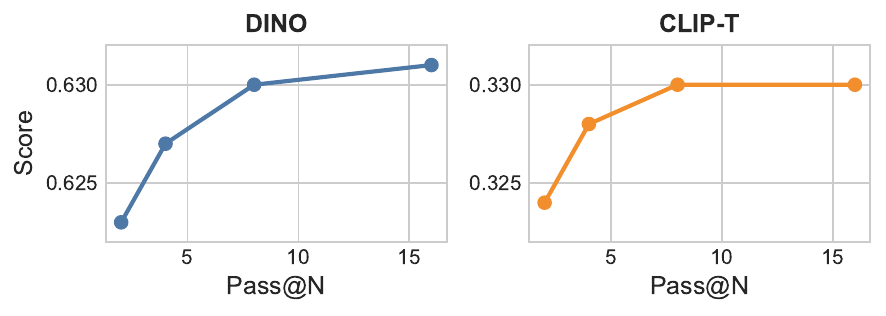}
    \vspace{-3mm}
    \caption{\small Effectiveness of test-time scaling using Semantics-Driven Spatial Layout Planning. }
    \label{fig:test_time}
\end{figure}
Our semantics-driven spatial layout planning mechanism enables effective test-time scaling to enhance performance (Section~\ref{sec:test_time_scaling}). This involves generating $N$ diverse candidate spatial plans and Vision CoT sequences, selecting the best image via CLIP similarity (Pass@N). Figure~\ref{fig:test_time} demonstrates this effect: DINO and CLIP-T scores consistently increase as Pass@N rises from 2 to 16 (\eg, DINO: 0.623 $\to$ 0.631, CLIP-T: 0.324 $\to$ 0.330). This confirms that generating more planning candidates and selecting the best improves image fidelity and text-alignment, providing a valuable mechanism for trading computation for quality at test time.

\subsection{Qualitative Analyses}
Figure~\ref{fig:case} shows qualitative comparisons between \textbf{MUSIC}, UNO~\cite{wu2025less}, and OmniGen~\cite{xiao2024omnigen} on multi-subject in-context generation. Each example includes reference subjects, a text prompt, and the corresponding generated images. \textbf{MUSIC} synthesizes complex scenes more coherently, accurately preserving subject identities and spatial relationships (e.g., objects in the living room or multiple interacting subjects on the beach). While baselines produce reasonable results, \textbf{MUSIC} achieves higher fidelity, more natural layouts, and better semantic consistency, consistent with  quantitative and human evaluations.

\section{Conclusion}
Generating images with multiple specific subjects remains challenging, often leading to missing subjects and semantic drift. We present MUSIC, the first MLLM for \textbf{MU}lti-\textbf{S}ubject \textbf{I}n-\textbf{C}ontext image generation. To address data scarcity, we design an automatic scalable data pipeline and enhance multi-subject reasoning through a vision chain-of-thought and semantics-driven spatial layout planning, enabling effective test-time scaling. Training on complex subject compositions further strengthens reasoning ability. We also introduce MSIC, a new benchmark for multi-subject in-context generation. Experiments show that MUSIC substantially surpasses other methods in identity fidelity, semantic consistency, and overall image quality.

\section*{Limitations}
Although MUSIC significantly outperforms baselines in handling multiple subjects and shows better scaling properties, the performance, as demonstrated in our experiments, still exhibits a degradation trend as the number of subjects increases.
Moreover, our test-time scaling via Pass@N improves quality at the cost of linearly increasing inference time, limiting applicability in low-latency scenarios. Future work could explore more efficient plan generation or alternative selection mechanisms.

\bibliography{ref}
\clearpage
\appendix

\section{Prompt}\label{app:prompt}
\begin{tcolorbox}[
colback=white,
colframe=black,
arc=1mm, 
auto outer arc,
title={Caption Generation},
breakable
]\small

Create a scene for an image generation model by semantically choosing at least half of these objects: \{classes\_str\} to fit a coherent setting. Use any style. Describe the scene in one concise English paragraph. no think.
\end{tcolorbox}

\begin{tcolorbox}[
colback=white,
colframe=black,
arc=1mm, 
auto outer arc,
title={Object Filtering},
breakable
]\small
Image: \{Image\_with\_box\_label\}


Text: 

\hangindent=5mm \hangafter=0 I have a scene description to evaluate. Here's the info:

\hangindent=5mm \hangafter=0 - Description: \{caption\}

\hangindent=5mm \hangafter=0 Think these questions in mind to check the description:

\hangindent=5mm \hangafter=0 1. Does the description avoid excessive adjectives or storytelling? (No more than 2 adjectives, no narrative.)

\hangindent=5mm \hangafter=0 2. Are all classes (\{', '.join(classes\_str)\}) mentioned in the description?

\hangindent=5mm \hangafter=0 Output:

\hangindent=5mm \hangafter=0 - If any answer is 'No', list violations like ['Missing class: xx'] and end with 'Please revise.'

\hangindent=5mm \hangafter=0 - If all answers are 'Yes', say 'Meets all criteria.'

\end{tcolorbox}

\begin{tcolorbox}[
colback=white,
colframe=black,
arc=1mm, 
auto outer arc,
title={Subject Image Transformation (Simple)},
breakable
]\small
Image: \{Image\_cropped\}


Text:

\hangindent=5mm \hangafter=0 A \{class\_name\} viewed from a different perspective, maintaining its core features and details.
        
\end{tcolorbox}

\begin{tcolorbox}[
colback=white,
colframe=black,
arc=1mm, 
auto outer arc,
title={Subject Image Transformation (Complex)},
breakable
]\small
Image: \{Image\_cropped\}


Text:

\hangindent=5mm \hangafter=0 i) (\textit{Random Optional}) \textit{With one randomly sampled object:}

\hangindent=5mm \hangafter=0 A cozy scene shows a \{class\_name\} sitting close to a \{random\_class\}, with their positions suggesting a natural interaction. The environment is filled with random objects, and neither item dominates the scene.

\hangindent=5mm \hangafter=0 ii) (\textit{Random Optional}) \textit{With two randomly sampled objects:}

\hangindent=5mm \hangafter=0 The \{class\_name\}, partially out of focus, lies near the camera; a \{random\_class\_1\} stands in full view, and a \{random\_class\_2\} is seen beyond a pile of scattered objects.

\hangindent=5mm \hangafter=0 iii) (\textit{Random Optional}) \textit{With three randomly sampled objects:}

\hangindent=5mm \hangafter=0 In a sun-drenched garden, the \{class\_name\} leans gently against a wooden crate. A \{random\_class\_1\} is placed on the grass nearby, with a \{random\_class\_2\} resting atop the crate. A \{random\_class\_3\} hangs lazily from a tree branch, swaying slightly with the breeze.

\end{tcolorbox}

\begin{tcolorbox}[
colback=white,
colframe=black,
arc=1mm, 
auto outer arc,
title={Simulated User Instruction Generation},
breakable
]\small
i) (\textit{Optional for Training}) \textit{With subject ID:}


Image: \{Image\_with\_box\_label\}


Text: 

\hangindent=5mm \hangafter=0 Create a scene description for image annotation using \{classes\_str\}

\hangindent=5mm \hangafter=0 Use short phrases, and describe how these objects might relate or be arranged in a scene. Clearly refer to each object with its source image (\eg, cat from image 1). Allow slight ambiguity, and mention each given object at least once relative to another or a landmark. Keep it concise, 5-30 words.

\hangindent=5mm \hangafter=0 Tempelate:

\hangindent=5mm \hangafter=0 The dog from image 0 rolls around in the grass while the cat from image 1 watches from a rock. A tall tree from image 2 casts shadows across the scene.


ii) (\textit{Optional for Training}) \textit{Without subject ID:}


Image: \{Image\}


Text:

\hangindent=5mm \hangafter=0 Create a detailed prompt for a generative model to create an image, depicting a harmonious scene composed of the following objects: \{classes\_str\}.
Describe how these objects interact or are arranged in the scene to form a cohesive and visually appealing composition.

\end{tcolorbox}

\begin{tcolorbox}[
colback=white,
colframe=black,
arc=1mm, 
auto outer arc,
title={CoT Generation},
breakable
]\small
       
Image: \{Image\_with\_box\_label\}


Text: 

\hangindent=5mm \hangafter=0 You are an expert in multi-image scene composition and visual grounding. Given initial prompt with subjects from different sub-image, create a detailed composition, step-by-step reasoning (CoT) to describe the entire scene in given target image.

\hangindent=5mm \hangafter=0 \#\#\# Input
\hangindent=5mm \hangafter=0 - **Initial Prompt**: "\{initial\_prompt\}"

\hangindent=5mm \hangafter=0 \#\#\# Note
\hangindent=5mm \hangafter=0 1. Clearly describe spatial relationships and interactions between objects (use terms like beside, behind, above, near, aligned with, etc.).

\hangindent=5mm \hangafter=0 2. Maintain a clear and logical flow, progressively describing the background, foreground, object positions, and their visual relationships.

\hangindent=5mm \hangafter=0 3. Include image IDs when needed to distinguish between multiple objects of the same category.

\hangindent=5mm \hangafter=0 4. Language: English. Provide at least 300 words.  
\end{tcolorbox}

\section{Data Example}\label{app:app}
\subsubsection*{Data Example 1:}
\begin{tcolorbox}[
colback=white,
colframe=black,
arc=1mm, 
auto outer arc,
title={Sampled Classes},
breakable
]\small
"toaster", "lamp", "plate", "potted plant", "chair couch", "radiator", "book", "sneakers", "sneakers leather shoes", "orange"
\end{tcolorbox}

\begin{tcolorbox}[
colback=white,
colframe=black,
arc=1mm, 
auto outer arc,
title={Detected Classes},
breakable
]\small
"microwave", "desk", "potted plant", "desk cabinet", "trolley", "vase", "carpet", "carpet"
\end{tcolorbox}

\begin{tcolorbox}[
colback=white,
colframe=black,
arc=1mm, 
auto outer arc,
title={Caption},
breakable
]\small
In the cozy living room, the lamp from image 1 casts a warm glow over the table where the toaster from image 0 sits, next to the plate from image 2. The potted plant from image 3 adds a touch of greenery, while the chair couch from image 4 provides a comfortable spot for reading. The radiator from image 5 ensures warmth, and the book from image 6 rests nearby. Sneakers from image 7 and sneakers leather shoes from image 8 hint at someone's presence, and an orange from image 9 adds a splash of color to the scene.
        
\end{tcolorbox}

\begin{tcolorbox}[
colback=white,
colframe=black,
arc=1mm, 
auto outer arc,
title={Simulated User Instruction},
breakable
]\small
The microwave from image 0 sits on the desk from image 1, which is covered by a green carpet from image 6. Nearby, a potted plant from image 2 adds greenery, while a desk cabinet from image 3 provides storage. A trolley from image 4 stands nearby, and a vase from image 5 adds color to the scene. Carpet from image 7 also appears in the scene.     
\end{tcolorbox}
\begin{tcolorbox}[
colback=white,
colframe=black,
arc=1mm, 
auto outer arc,
title={CoT},
breakable
]\small
\#\#\# Detailed Composition and Spatial Relationships\textbackslash n\textbackslash n\#\#\#\# Background:\textbackslash nThe scene opens with a cozy backyard setting, featuring a house with a red and white striped awning over the entrance. The house has a wooden deck leading up to the front door, which is partially visible through the awning. The surrounding area is lush with greenery, including bushes and trees, creating a serene and inviting atmosphere. This backdrop sets the stage for the rest of the elements in the scene.\textbackslash n\textbackslash n\#\#\#\# Foreground:\textbackslash nIn the foreground, there is a well-maintained garden area with various plants and flowers. A large, vibrant purple cabbage is prominently displayed in a terracotta pot, adding a pop of color and texture to the scene. Nearby, a small gray cat is sitting on the grass, seemingly relaxed and enjoying the peaceful environment. A few red flowers are scattered around the garden, enhancing the natural beauty of the space.\textbackslash n\textbackslash n\#\#\#\# Desk and Microwave:\textbackslash nMoving forward, we see a wooden desk placed on the patio. On this desk, there is a microwave from image 0, which is positioned centrally. The desk itself is covered by a green carpet from image 7, providing a soft and comfortable surface for the microwave and other items. The desk cabinet from image 3 is situated behind the desk, offering additional storage space. The desk cabinet is aligned with the desk, ensuring that it complements the overall setup without obstructing any view or functionality.\textbackslash n\textbackslash n\#\#\#\# Trolley and Vase:\textbackslash nTo the right of the desk, a blue trolley from image 4 is parked. The trolley is empty but appears ready for use, suggesting it might be used for transporting items within the garden or around the house. Adjacent to the trolley, a vase from image 5 is placed on the ground, adding a touch of elegance and color to the scene. The vase is filled with red flowers, which match the flowers in the garden, creating a cohesive and harmonious look.\textbackslash n\textbackslash n\#\#\#\# Potted Plant:\textbackslash nA potted plant from image 2 is positioned near the trolley, adding more greenery to the scene. The plant is placed on the ground, slightly away from the desk and trolley, allowing it to blend naturally into the environment. Its presence enhances the overall aesthetic of the backyard, making it feel more alive and vibrant.\textbackslash n\textbackslash n\#\#\#\# Overall Scene:\textbackslash nThe entire scene is a harmonious blend of indoor and outdoor elements, creating a functional yet aesthetically pleasing environment. The microwave on the desk, the green carpet, the desk cabinet, the trolley, the vase, and the potted plant all contribute to a well-organized and inviting space. The background of the house and the lush greenery provide a calming and natural backdrop, making the scene feel both practical and serene. The spatial relationships between these objects are carefully arranged to ensure that each element has its own place while still contributing to the overall harmony of the scene."

\end{tcolorbox}
\begin{tcolorbox}[
colback=white,
colframe=black,
arc=1mm, 
auto outer arc,
title={Spatial Layout Prompt},
breakable
]\small
Here is the segmentation map focusing on piano, person, flower, carpet, chair, stool, frame, desk cabinet, cabinet, blackboard:\textbackslash n<patch>[0] others [1] others [2] others [3] others [4] others [5] others [6] others [7] others [8] blackboard [9] others [10] others [11] others [12] others [13] others [14] others [15] others [16] blackboard [17] others [18] others [19] others [20] others [21] others [22] cabinet [23] cabinet [24] blackboard [25] frame [26] others [27] others [28] person [29] others [30] cabinet, flower [31] cabinet, flower [32] others [33] frame [34] others [35] person [36] person [37] chair [38] flower, piano [39] flower, piano [40] desk cabinet [41] desk cabinet [42] desk cabinet [43] person [44] chair, person [45] chair [46] piano [47] piano [48] desk cabinet [49] desk cabinet [50] desk cabinet [51] chair, person [52] carpet, chair [53] carpet, chair [54] carpet, piano [55] piano [56] desk cabinet, stool [57] desk cabinet [58] desk cabinet [59] carpet [60] carpet [61] carpet [62] carpet [63] others</patch>\textbackslash n Now, generate an image.
\end{tcolorbox}

\subsubsection*{Data Example 2:}
\begin{tcolorbox}[
colback=white,
colframe=black,
arc=1mm, 
auto outer arc,
title={Sampled Classes},
breakable
]\small
"ship", "sports car"
\end{tcolorbox}

\begin{tcolorbox}[
colback=white,
colframe=black,
arc=1mm, 
auto outer arc,
title={Detected Classes},
breakable
]\small
"ship", "sports car"
\end{tcolorbox}

\begin{tcolorbox}[
colback=white,
colframe=black,
arc=1mm, 
auto outer arc,
title={Caption},
breakable
]\small
A sleek red sports car speeds along a coastal highway, its shiny body reflecting the golden hues of a setting sun, while in the distance, a large ship cuts through the waves of the ocean, its massive silhouette contrasting with the car's nimble form, capturing the essence of speed and adventure against the vastness of the sea.
        
\end{tcolorbox}

\begin{tcolorbox}[
colback=white,
colframe=black,
arc=1mm, 
auto outer arc,
title={Simulated User Instruction},
breakable
]\small
Imagine the ship from image 0 sailing in the distance, with the sports car from image 1 parked on the road near the ocean, both under the warm glow of the setting sun.
                    
\end{tcolorbox}
\begin{tcolorbox}[
colback=white,
colframe=black,
arc=1mm, 
auto outer arc,
title={CoT},
breakable
]\small
\#\#\# Detailed Composition and Reasoning\textbackslash n\textbackslash nThe scene is a harmonious blend of two distinct elements: a ship from image 0 and a sports car from image 1, set against the backdrop of a breathtaking sunset over the ocean. The composition is carefully crafted to ensure that each element contributes to a cohesive and visually appealing narrative.\#\#\#\# Background:\textbackslash n- **Sunset**: The sky is painted in warm hues of orange and yellow, transitioning into a soft gradient as it meets the horizon. The sun is positioned low on the horizon, casting a golden glow across the entire scene. This creates a serene and tranquil atmosphere, perfect for the contemplative mood suggested by the ship and the dynamic energy of the sports car.\textbackslash n- **Ocean**: The calm waters of the ocean reflect the vibrant colors of the sunset, creating a shimmering effect that enhances the overall beauty of the scene. The waves gently lap against the rocky cliffs, adding a sense of movement and life to the otherwise still landscape.\textbackslash n- **Cliffs**: On the right side of the image, rugged cliffs rise steeply from the water's edge. Their dark, textured surfaces contrast sharply with the bright colors of the sunset, providing a dramatic and natural frame for the scene.\#\#\#\# Foreground:\textbackslash n- **Sports Car (Image 1)**: A sleek red sports car is positioned prominently in the foreground, driving along a coastal road. The car is angled slightly towards the viewer, suggesting motion and speed. Its polished exterior reflects the warm light of the sunset, adding depth and vibrancy to the image. The car's design is modern and aerodynamic, with sharp lines and curves that emphasize its high-performance capabilities.\textbackslash n- **Road**: The road on which the car is traveling is a two-lane highway with a metal guardrail running parallel to the cliffside. The road appears smooth and well-maintained, curving gently along the coastline. The guardrail adds a sense of safety and structure to the scene, while also emphasizing the proximity of the car to the edge of the cliff.\#\#\#\# Spatial Relationships:\textbackslash n- **Ship (Image 0)**: The ship is situated further out in the ocean, positioned behind the cliffs and slightly to the left of the sports car. It is a large cargo ship, identifiable by its size and the cranes visible on its deck. The ship is sailing away from the viewer, moving towards the horizon where it blends seamlessly with the sunset. Its presence adds a sense of scale and grandeur to the scene, contrasting with the smaller, more personal scale of the sports car.\textbackslash n- **Alignment and Interaction**: The sports car is positioned closer to the viewer, creating a sense of depth and perspective. The ship, being farther away, provides a sense of distance and vastness. The cliffs act as a natural barrier between the two elements, subtly separating them yet keeping them within the same frame. The warm light of the sunset unifies the scene, tying together the ship, the car, and the ocean into a single, cohesive composition.\#\#\#\# Visual Flow:\textbackslash nThe composition flows naturally from the foreground to the background. The sports car, with its vibrant red color and sleek design, draws the viewer's attention immediately. As the eye moves upward, the warm glow of the sunset and the reflective ocean create a sense of serenity and beauty. Finally, the ship in the distance adds a layer of intrigue and scale, completing the scene with a sense of adventure and exploration.In summary, the scene is a masterful blend of motion and tranquility, with the sports car symbolizing human ambition and the ship representing the vastness of the world. The sunset serves as a unifying element, creating a harmonious balance between the different components of the image.
\end{tcolorbox}
\begin{tcolorbox}[
colback=white,
colframe=black,
arc=1mm, 
auto outer arc,
title={Spatial Layout Prompt},
breakable
]\small
Here is the segmentation map focusing on ship, sports car:\textbackslash n<patch>[0] others [1] others [2] others [3] others [4] others [5] others [6] others [7] others [8] others [9] others [10] others [11] others [12] others [13] others [14] others [15] others [16] others [17] ship [18] ship [19] others [20] others [21] others [22] others [23] others [24] others [25] ship [26] ship [27] ship [28] others [29] others [30] others [31] others [32] others [33] sports car [34] sports car [35] sports car [36] sports car [37] sports car [38] others [39] others [40] sports car [41] sports car [42] sports car [43] sports car [44] sports car [45] sports car [46] sports car [47] others [48] others [49] others [50] others [51] others [52] sports car [53] sports car [54] sports car [55] others [56] others [57] others [58] others [59] others [60] others [61] others [62] others [63] others</patch>\textbackslash n Now, generate an image.
                    
\end{tcolorbox}

\subsubsection*{Data Example 3:}
\begin{tcolorbox}[
colback=white,
colframe=black,
arc=1mm, 
auto outer arc,
title={Sampled Classes},
breakable
]\small
"trumpet", "sushi", "flute", "tissue", "brush", "bracelet", "piano", "trophy", "coffee machine", "candy", "globe"
\end{tcolorbox}

\begin{tcolorbox}[
colback=white,
colframe=black,
arc=1mm, 
auto outer arc,
title={Detected Classes},
breakable
]\small
"piano", "person", "flower", "carpet", "chair", "stool", "frame", "desk cabinet", "cabinet", "blackboard"
\end{tcolorbox}

\begin{tcolorbox}[
colback=white,
colframe=black,
arc=1mm, 
auto outer arc,
title={Caption},
breakable
]\small
In a cozy, sunlit studio with floor-to-ceiling windows, a painter sits at an easel, a brush in hand, surrounded by scattered art supplies and a small globe on a nearby shelf, symbolizing inspiration from around the world. To their left, a coffee machine steams quietly, while a tray of sushi and a candy dish sit on the desk, offering a delightful break. Nearby, a trumpet and a flute rest on a piano, suggesting a dual passion for music and art. A trophy sits on the windowsill, reflecting the artist's past achievements, while a tissue and a bracelet lie nearby, hinting at a moment of pause or reflection.
        
\end{tcolorbox}

\begin{tcolorbox}[
colback=white,
colframe=black,
arc=1mm, 
auto outer arc,
title={Simulated User Instruction},
breakable
]\small
The person from image 1 sits on a chair from image 4, painting near a piano from image 0. A flower from image 2 adorns the room, and a carpet from image 3 covers the floor. The blackboard from image 9 is visible in the background, and a desk cabinet from image 7 is nearby. Stool from image 5, frame from image 6, cabinet from image 8 also appear in the scene.
                    
\end{tcolorbox}
\begin{tcolorbox}[
colback=white,
colframe=black,
arc=1mm, 
auto outer arc,
title={CoT},
breakable
]\small
\#\#\# Detailed Composition and Spatial Relationships\textbackslash n\textbackslash n\#\#\#\# Background:\textbackslash nThe scene is set in a cozy, well-lit studio with large windows that offer a serene view of lush greenery outside. The natural light streaming through the windows illuminates the space, creating a warm and inviting atmosphere. In the background, a blackboard from Image 9 is visible, suggesting this might be a creative or educational workspace. The blackboard is positioned against the wall, slightly to the left of the center of the room, providing a focal point for any notes or drawings.\textbackslash n\textbackslash n\#\#\#\# Foreground:\textbackslash nIn the foreground, a person is seated on a wooden chair from Image 4, engrossed in painting on an easel. The individual appears focused, holding a paintbrush in one hand while the other rests on the canvas. The easel is placed on a desk cabinet from Image 7, which is situated to the left of the person. The desk cabinet has several drawers and compartments, offering storage for art supplies and other materials. Adjacent to the desk cabinet, a stool from Image 5 can be seen, likely used by the artist for additional seating or to rest their feet during long sessions.\textbackslash n\textbackslash n\#\#\#\# Objects and Their Interactions:\textbackslash n- **Piano from Image 0**: Positioned to the right of the person, a grand piano adds a touch of elegance and suggests a multifunctional use of the space. The piano's polished surface reflects some of the ambient light, enhancing its prominence in the room.\textbackslash n- **Flower from Image 2**: A vase of flowers is placed on a small table near the piano, adding a splash of color and life to the scene. The flowers are positioned slightly behind the piano, creating a harmonious balance between the musical instrument and the floral arrangement.\textbackslash n- **Carpet from Image 3**: A soft pink carpet lies on the wooden floor, extending from the center of the room towards the piano. The carpet not only adds warmth to the space but also provides a comfortable area for the artist to sit or move around.\textbackslash n- **Desk Cabinet from Image 7**: As mentioned earlier, the desk cabinet serves as a functional piece of furniture, housing various items such as art supplies and personal belongings. It is aligned with the desk where the easel is placed, facilitating easy access to tools and materials.\textbackslash n- **Frame from Image 6**: A framed picture or artwork hangs on the wall above the desk cabinet, adding an element of personalization and aesthetic appeal to the room. The frame is positioned slightly above and to the left of the desk, drawing the viewer's eye upwards.\textbackslash n- **Stool from Image 5**: The stool is located to the left of the desk cabinet, providing an alternative seating option for the artist. Its placement ensures that the artist can easily switch between sitting on the chair and using the stool without disrupting their workflow.\textbackslash n\textbackslash n\#\#\#\# Overall Flow:\textbackslash nThe scene is meticulously arranged to create a harmonious and productive environment. The artist is comfortably seated, surrounded by essential tools and personal touches that enhance both functionality and aesthetics. The natural light from the windows complements the warm tones of the wooden furniture and the soft pink carpet, creating a tranquil and inspiring workspace. The presence of the piano and the vase of flowers adds layers of depth and character, making the studio feel like a sanctuary for creativity and relaxation.
\end{tcolorbox}
\begin{tcolorbox}[
colback=white,
colframe=black,
arc=1mm, 
auto outer arc,
title={Spatial Layout Prompt},
breakable
]\small
Here is the segmentation map focusing on piano, person, flower, carpet, chair, stool, frame, desk cabinet, cabinet, blackboard:\textbackslash n<patch>[0] others [1] others [2] others [3] others [4] others [5] others [6] others [7] others [8] blackboard [9] others [10] others [11] others [12] others [13] others [14] others [15] others [16] blackboard [17] others [18] others [19] others [20] others [21] others [22] cabinet [23] cabinet [24] blackboard [25] frame [26] others [27] others [28] person [29] others [30] cabinet, flower [31] cabinet, flower [32] others [33] frame [34] others [35] person [36] person [37] chair [38] flower, piano [39] flower, piano [40] desk cabinet [41] desk cabinet [42] desk cabinet [43] person [44] chair, person [45] chair [46] piano [47] piano [48] desk cabinet [49] desk cabinet [50] desk cabinet [51] chair, person [52] carpet, chair [53] carpet, chair [54] carpet, piano [55] piano [56] desk cabinet, stool [57] desk cabinet [58] desk cabinet [59] carpet [60] carpet [61] carpet [62] carpet [63] others</patch>\textbackslash n Now, generate an image.
                    
\end{tcolorbox}

\begin{table*}[!t]\small
\centering
\begin{tabular}{lccc}
\toprule
\textbf{Method} & \textbf{DINO} $\uparrow$ & \textbf{CLIP-I} $\uparrow$ & \textbf{CLIP-T} $\uparrow$ \\
\midrule
SDXL + IP-Adapter & 0.571 & 0.738 & 0.294 \\
PixArt-$\Sigma$ + IP-Adapter & 0.583 & 0.746 & 0.298 \\
DiT-XL + Subject Adapter & 0.589 & 0.751 & 0.301 \\
UNO (FLUX-1.0-DEV) & 0.541 & 0.721 & 0.296 \\
\midrule
MUSIC (Ours) & \textbf{0.622} & \textbf{0.812} & \textbf{0.322} \\
\bottomrule
\end{tabular}
\caption{Comparison with strong diffusion backbones adapted for subject-driven generation on MSIC.}
\label{tab:diffusion_comparison}
\end{table*}
\begin{table*}[!t]\small
\centering
\begin{tabular}{lcc}
\toprule
\textbf{Pipeline Stage} & \textbf{Failure Rate (\%)} & \textbf{Mitigation Strategy} \\
\midrule
T2I semantic mismatch & 9.8 & Prompt filtering and regeneration \\
OVD miss / false detection & 14.6 & Area threshold + VLM verification \\
VLM hallucinated validation & 6.3 & ID-string consistency check \\
SAM2 incorrect mask & 11.2 & CLIP-based mask vs. box selection \\
\midrule
Final retained samples & \multicolumn{2}{c}{\textbf{68.1\%}} \\
\bottomrule
\end{tabular}
\caption{Per-stage error rates and mitigation strategies in the data construction pipeline.}
\label{tab:pipeline_errors}
\end{table*}

\section{Complex Subject Images Data Generation}\label{app:complex}

Current subject-to-image generation tasks typically focus on generating scenes based on a few subject images corresponding to target objects. However, in real-world applications, users may want to select specific objects from complex scene images and generate new scenes based on those selections. Our method, leveraging \textit{in-context learning}, effectively addresses this more practical and challenging scenario.

To support this, we generate a dedicated dataset where subject objects are embedded into complex scenes and transformed accordingly. Since conventional I2I models (\eg, UNO-FLUX) lack reasoning capabilities, we first need to construct rich and contextually meaningful prompts.

Specifically, we traverse our previously generated dataset to compile a comprehensive list of object categories. This list is then provided to GPT-4o~\cite{hurst2024gpt}, to automatically generate a dictionary mapping each category to a set of semantically related categories. 

\textbf{For example:} 
\begin{itemize}
    \item \textbf{Key}: \textit{Chair} $\rightarrow$ \textbf{Value}: \textit{Stool, Bench, Couch}
    \item \textbf{Key}: \textit{Bottle} $\rightarrow$ \textbf{Value}: \textit{Cup, Glass, Jar}
\end{itemize}

Using this dictionary, we instruct GPT-4o to produce diverse prompt templates that capture natural object interactions and varied scene compositions. Examples include:

\begin{itemize}
    \item \textit{A cozy scene shows a \{class\_name\} sitting close to a \{random\_class\}, with their positions suggesting a natural interaction. The environment is filled with random objects, and neither item dominates the scene.}
    
    \item \textit{The \{class\_name\}, partially out of focus, lies near the camera; a \{random\_class\_1\} stands in full view, and a \{random\_class\_2\} is seen beyond a pile of scattered objects.}
    
    \item \textit{In a sun-drenched garden, the \{class\_name\} leans gently against a wooden crate. A \{random\_class\_1\} is placed on the grass nearby, with a \{random\_class\_2\} resting atop the crate. A \{random\_class\_3\} hangs lazily from a tree branch, swaying slightly with the breeze.}
\end{itemize}

When embedding subject images into complex scenes, we randomly select both a prompt template and object categories to construct a detailed prompt. This prompt is then provided to the T2I model to generate complex subject cases.

\section{Additional Experimental Analysis and Validation}
\label{app:additional}

\subsection{Comparison with Stronger Diffusion Backbones}
\label{app:diffusion_comparison}
To better position MUSIC within the current state-of-the-art landscape, we provide additional comparisons with strong diffusion-based backbones adapted for subject-driven generation.
While many recent diffusion transformers do not natively support multi-subject in-context conditioning with reference images, we equip them with subject adapters for a controlled and fair comparison.
All models are evaluated on the MSIC benchmark using identical prompts and reference subjects.
As shown in Table~\ref{tab:diffusion_comparison}, despite stronger diffusion backbones improving base image fidelity, they still exhibit subject omission and identity entanglement as the number of subjects increases.
MUSIC consistently outperforms these methods due to its explicit reasoning and planning formulation.

\subsection{Error Propagation Analysis of the Automated Data Pipeline}
\label{app:pipeline_analysis}
Our automated data construction pipeline integrates multiple pretrained foundation models, each of which may introduce noise.
To quantify cumulative error propagation, we measure per-stage failure rates on a random subset of generated samples and summarize the mitigation strategies used at each stage (Table~\ref{tab:pipeline_errors}).
Although individual stages introduce moderate noise, cascading filters significantly reduce error accumulation.
Empirically, models trained on this automatically curated dataset achieve consistent gains, indicating robustness to realistic supervision noise.

\begin{table}[!t]\small
\centering
\setlength{\tabcolsep}{2pt}
\resizebox{\linewidth}{!}{
\begin{tabular}{lcc}
\toprule
\textbf{Method} & \textbf{Patch IoU} $\uparrow$ & \textbf{Category Coverage} $\uparrow$ \\
\midrule
Random layout & 0.21 & 0.48 \\
UNO (implicit layout) & 0.34 & 0.61 \\
MUSIC (predicted layout) & \textbf{0.47} & \textbf{0.76} \\
\bottomrule
\end{tabular}}
\caption{Agreement between predicted spatial layouts and segmentation masks.}
\label{tab:layout_agreement}
\end{table}
\begin{table}[!t]\small
\centering
\begin{tabular}{lcc}
\toprule
\textbf{Method} & \textbf{Relation Accuracy} $\uparrow$ & \textbf{Identity Fidelity} $\uparrow$ \\
\midrule
UNO & 0.63 & 0.72 \\
MUSIC & \textbf{0.74} & \textbf{0.81} \\
\bottomrule
\end{tabular}
\caption{Performance on relative spatial relation prompts on MSIC.}
\label{tab:cot_relations}
\end{table}

\subsection{Validation of the Learned Spatial Layout Prior}
\label{app:layout_validation}
The semantics-driven spatial layout in MUSIC serves as a high-level prior rather than pixel-accurate supervision.
To verify that MUSIC learns meaningful spatial structure rather than memorizing noisy layouts, we measure the agreement between predicted layouts and SAM2 segmentation masks on held-out images.
As reported in Table~\ref{tab:layout_agreement}, the learned spatial layout prior aligns well with object spatial distributions and is not dominated by hallucinated patterns.

\subsection{Effectiveness of Vision Chain-of-Thought for Spatial Relations}
\label{app:cot_relations}
Vision Chain-of-Thought (CoT) in MUSIC focuses on relative spatial relations (e.g., left, right, behind), which are more robust and natural for complex multi-subject composition.
We evaluate relational accuracy on prompts requiring relative spatial reasoning (Table~\ref{tab:cot_relations}).

\end{document}